# Design and Development of a Software System for Swarm Intelligence Based Research Studies

*Utku Kose*
Usak University
Uşak Üniversitesi Rektörlüğü 1 Eylül Kampüsü  64200, Uşak, Turkey
Phone: :+90.276.221 21 21 (pbx)
utku.kose@usak.edu.tr

**Abstract**
This paper introduce a software system including widely-used Swarm Intelligence algorithms or approaches to be used for the related scientific research studies associated with the subject area. The programmatic infrastructure of the system allows working on a fast, easy-to-use, interactive platform to perform Swarm Intelligence based studies in a more effective, efficient and accurate way. In this sense, the system employs all of the necessary controls for the algorithms and it ensures an interactive platform on which computer users can perform studies on a wide spectrum of solution approaches associated with simple and also more advanced problems.

## 1. Introduction

Today, the Swarm Intelligence (SI) is a remarkable and important area of the Artificial Intelligence field in the context of the Computer Science. The research scope of this area is associated with the problem-solution approaches on employing the collective behavior of natural or artificial self-organized structures or systems. In this sense, many different types of SI based algorithms have been introduced to provide better solutions for especially real-world based problems. The SI area and its related working scope is very active and newer and more advanced algorithms aiming to provide better approaches rather than the current ones are still designed and developed in a rapid way. As being parallel with the mentioned efforts, there is also be a remarkable need for especially researchers to benefit from tools, programs or general software systems providing solution approaches of the SI based algorithms.

Objective of this paper is to introduce a software system, which employs widely-used Swarm Intelligence algorithms or approaches on a common working environment to be used for the related scientific research studies associated with the subject area. Basically, the system has been designed and developed to provide a typical form-based software environment for the popular and newly developed Swarm Intelligence based algorithms in the mentioned scope. The programmatic infrastructure of the system enables computer users to work on a fast, easy-to-use, interactive platform to perform their studies in a more effective, efficient and accurate way. In this sense, the system employs all of the necessary controls for the algorithms provided and it ensures an interactive platform on which computer users can perform studies on a wide spectrum of solution approaches associated with many types of problems changing from simple to more advanced ones. The author thinks that this software system is a general 'tool box' to be used for SI based research approaches.

The rest of the paper is organized as follows: In the second section, foundations of the software system are described briefly. At this point, some essential references are listed for readers to direct them to have better idea about some SI based algorithms which are included within the designed and developed software system. Next, general design structure and the most important using features and functions of the software system are introduced – explained in the third section. After the third section, a survey work performed among some researchers to have better idea about effectiveness and success of the software system is explained briefly in the fourth section. Finally, results of the realized work and also some future works related to the software system are discussed in the last section.

## 2. Foundations

The software system designed and developed in this work comes with some popular SI based algorithms to allow working via different kinds of solution based approaches for solving the related problems. In this sense, Particle Swarm Optimization (PSO), Ant Colony Optimization (ACO), Artificial Bee Colony (ABC), Intelligent Water Drops (IWDs) Algorithm and Firefly Algorithm (FA) are included within the software platform. The algorithms have been chosen according to their popularities and / or interesting features in the context of problem solving approaches. In order not to affect flow of the paper and also its readability, readers are directed to some essential references to have more idea about the mentioned algorithms. The related references are listed as below:

- PSO is a global optimization algorithm, which have been introduced by Kennedy and Eberhart [1]. It is based on the simulation of social behaviors appeared in a fish school or bird flock. More information about the PSO can be found in [2 – 4].

- ACO is a SI based algorithm, which have been introduced by Dorigo [5]. Objective of this algorithm is to search for an optimal path in a graph structure by simulating the behavior of ants seeking an appropriate path between the colony and a food source. More information about the ACO can be found in [6 – 8].

- ABC is an optimization algorithm, which have been introduced by Karaboga [9]. Approach of the ABC algorithm is based on the intelligent foraging behavior of a honey-bee swarm. More information about the ABC can be found in [10 – 12].

- IWDs Algorithm is a SI based algorithm, which have been introduced by Shah-Hosseini [13]. Generally, it is similar to the ACO but its working mechanism is based on dynamic of a natural river system. More information about this algorithm can be found in [14 – 16].

- FA is a SI based algorithm, which have been introduced by Yang [17]. Approach of this algorithm is based on flashing behaviors of fireflies. More information about the algorithm can be obtained from [18 – 20].

## 3. Design and development of a software system for swarm intelligence based research studies

The software system, which has been designed and developed within this study, employs simple and visually improved, form-based controls to ensure the related objectives of the study. At this point; in order to have better idea about design of the system, it is better to examine the programmatic infrastructure of the system before explaining using features and functions.

### *3.1. Programmatic infrastructure*

Basically, the programmatic infrastructure of the software system is based on separate application codes combined within a single software system platform. In this sense, the whole system has been coded over the Microsoft Visual Studio 2010 platform by using the C# programming language. As mentioned before, the system is based on a typical form-based software approach and each application form associated with a specific SI based algorithm includes its own code structure as being independent from other ones and being connected with the main software system interface. Briefly, this infrastructure can be represented in a schema as shown in Fig. 1.

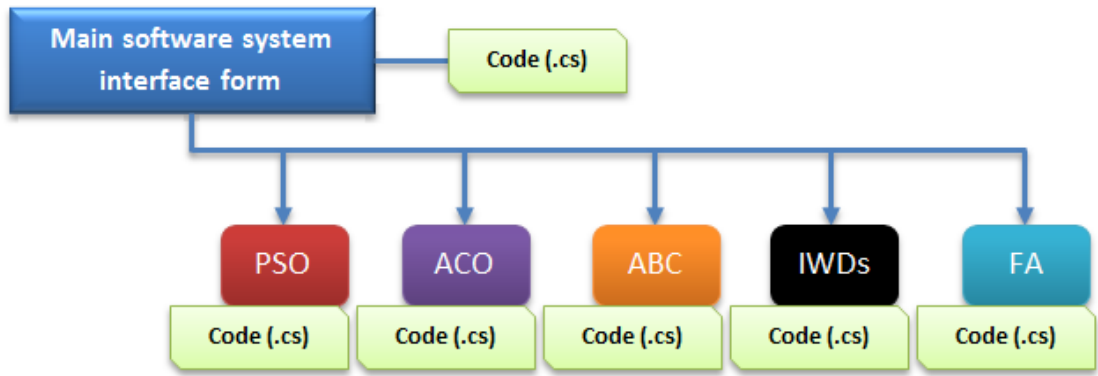

Figure 1. Programmatic infrastructure of the software system

### 3.2. Using features and functions

Generally, the usage of the developed software system is mostly based on typical form-based controls in order to provide an easy-to-use, fast software environment to improve using experiences positively. In this sense, each algorithm employed in the software system can be reached by using the provided button controls over the main software system interface (form – window). After clicking on the related button, corresponding interface – window of the chosen SI based algorithm is viewed. On a typical algorithm interface – window, it is possible to perform the related operations listed below:

- Defining a new problem and new parameters to perform a solution task associated with the problem scope of the algorithm.

- Opening a saved or predefined problem file to perform newer solution based tasks according to different parameters or structure combinations related to the algorithm (In order to ensure a fast and accurate working mechanism, 'problem files' are saved in the XML file format).

- Performing predefined problems that can be solved by only the related algorithm.

All of the mentioned operations can be performed easily by using the provided controls over the related interfaces – windows of each algorithm. It is also important that each algorithm interface is supported by visual controls to view obtained results with typical iteration-based graphics or problem oriented visual elements. For instance, resulting graph structures are automatically shown by the algorithm interfaces after solving some specific, popular problems like Travelling Salesman Problem (TSP), Vehicle Routing Problem (VCP)…etc. Visually improved using features and functions of the software system are critical aspects to provide more effective and useful platform to perform SI based research studies better.

Related to the designed and developed software system, some screenshots from the software system [interfaces of two algorithms (IWDs and ABC)] are represented in Fig. 2.

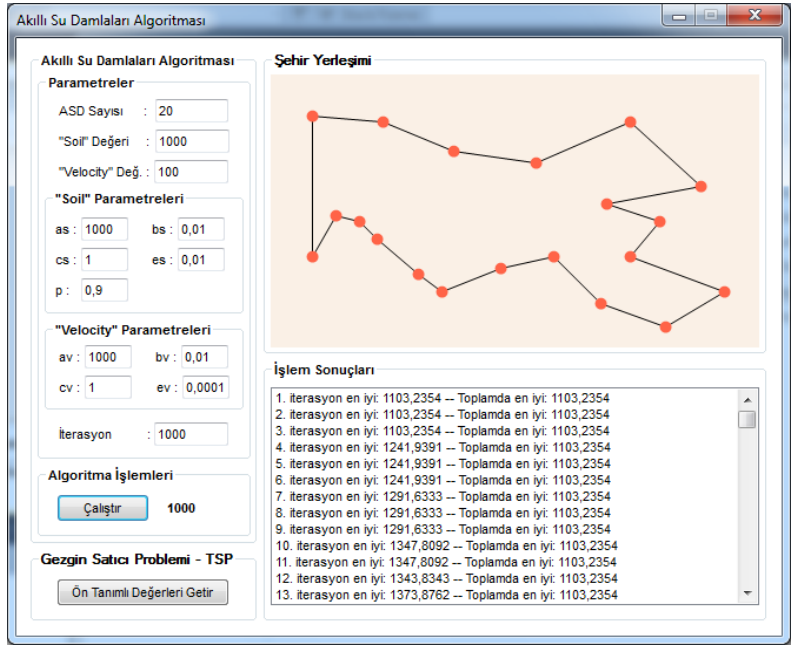

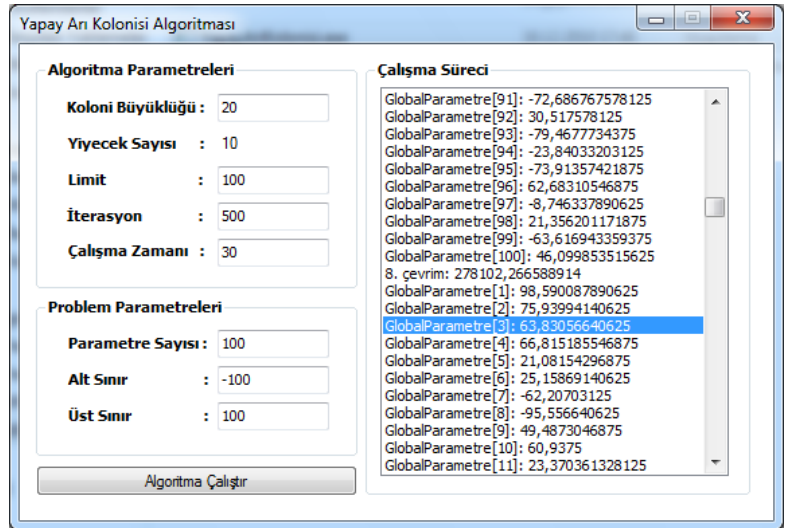

Figure 2. Some screenshots from the software system

## 4. Evaluation

In order to have more idea about effectiveness and success of the software system, it has been used by a total of 50 different scientists - researchers for two weeks and after the using period, the related scientists - researchers have filled a survey including some statements about the usage and effectiveness of the system. The related survey includes a total of 10 statements that can be evaluated via Likert scale. In this sense, scientists – researchers have typed '1' if they strongly disagree, '2' if they disagree, '3' if they have no clear opinion, '4' if they agree and '5' if they strongly agree with a given statement. The related results obtained with the survey study are shown in Table 1.

Table 1. The survey – evaluation results

| Statement | Number of responses for: | | | | |
|---|---|---|---|---|---|
| | 1 | 2 | 3 | 4 | 5 |
| This software system provides a fast research study based platform. | 0 | 0 | 3 | 8 | 39 |
| By using this system, it is easier to perform SI based research studies. | 0 | 0 | 0 | 7 | 43 |
| There must be more SI based algorithms within the software system. | 0 | 0 | 0 | 6 | 44 |
| I didn't like the general characteristic of this software system. | 37 | 10 | 3 | 0 | 0 |
| I liked the scope of the predefined problems for the related algorithms. | 0 | 0 | 1 | 6 | 43 |
| It is hard to perform SI based research studies via this software system. | 40 | 6 | 4 | 0 | 0 |
| I liked the visual controls provided on this software system. | 0 | 0 | 0 | 1 | 49 |
| I don't want to take part in this kind of study process again. | 45 | 4 | 1 | 0 | 0 |
| The system provides an effective approach for SI based res. studies. | 0 | 1 | 3 | 6 | 41 |
| The algorithms provided within the system provide accurate results. | 0 | 0 | 1 | 7 | 42 |
| | | | | **Total Respondents:** 50 | |

Results obtained with the survey study show that the software system introduced in this paper is evaluated as an effective and successful platform - environment that can be easily used for performing SI based research studies.

### 5. Conclusions and future work

The software system introduced in this paper ensures an effective platform, which employs widely-used SI based algorithms on a common working environment to be used for performing related scientific research studies. The system employs fast, simple, easy-to-use controls and interfaces to enable scientists – researchers to perform their works in a more effective, efficient and accurate way. In this sense, the system also provides visually improved controls to improve general using experience. Furthermore, the system also comes with predefined problem files that have been specially prepared for the widely-used algorithms included within the system.

According to the evaluation results, which have been obtained with a simple survey study, all of these mentioned advantages of the system were also confirmed and as general, the software system is evaluated as an effective and successful platform - environment that can be easily used for performing SI based research studies.

In addition to the expressed positive feedbacks, some evaluation results (survey responses) have also encouraged the author to plan some future works. For instance, number of algorithms employed within this system will be increased in future versions by adding different kinds of SI based algorithms. Additionally, future versions of the system will also enable scientists – researchers to reach to the system environment over the Web platform. Finally, there will probably be more visual and interactive controls within the future developments of the software system.